%% file: main.tex
\documentclass[letterpaper, 10 pt, conference]{ieeeconf}
\IEEEoverridecommandlockouts
\input{preamble}
\usepackage[T1]{fontenc}
\usepackage{caption}
\title{\LARGE \bf \textsc{Noctif3R}: Feed-Forward Monocular Real-Time SLAM\\
for Photon-Limited Scenes on Embedded Hardware}

\author{Mihir Chauhan$^{1}$, Aditya Uday Abhang$^{1}$, Kevin Biju Mathew$^{1}$, Aniket Bera$^{1}$%
\thanks{\raggedright $^{1}$Mihir Chauhan, Aditya Uday Abhang, Kevin Biju Mathew, and Aniket Bera are with the IDEAS Lab, Department of Computer Science, Purdue University, West Lafayette, IN, USA. %
        {\tt\small \{\href{mailto:chauhanm@purdue.edu}{chauhanm}\allowbreak,\href{mailto:aniketbera@purdue.edu}{aniketbera}\allowbreak\}@\allowbreak purdue\allowbreak.edu}}
}

\begin{document}
\maketitle

\thispagestyle{empty}
\pagestyle{empty}

\begin{abstract}

Robots carrying out tasks in dark environments need to localize from a single RGB camera, in light so low that the per-pixel signal approaches the sensor's own noise, on a power-constrained onboard computer, in real time. Each of these constraints has matured pipelines, but the intersection does not. Offline low-light reconstruction now recovers structure below $-4$~dB but is far too slow to run in real time, while the real-time monocular systems a robot can actually carry (DROID-SLAM, DPV-SLAM, VGGT-SLAM, $\pi^3$, CUT3R, EC3R-SLAM, ORB-SLAM3 and DSO) degrade or fail when SNR gets low. We measured how they fail: across the nine lowest darkness levels of our scenes, DROID-SLAM returns a full-length trajectory carrying no information about the camera's motion on all nine, VGGT-SLAM and CUT3R on eight, $\pi^3$ on seven, and DPV-SLAM on four. We present \sys, a monocular pipeline built on a low-light feed-forward pointmap front end with an explicit match gate, which returns three tracked trajectories and no uninformative ones, at the lowest error of any method where it tracks ($24$--$47\%$ of the no-information ceiling against $56$--$73\%$ for the strongest baseline), and at the narrowest coverage. On a real robot video take in which $86.5\%$ of delivered frames are entirely black, every configuration of ours stops after the lit beginning, while DROID-SLAM and DPV-SLAM each emit a pose for all $1178$ frames. Our method contribution is an embedded execution path for the Jetson AGX Orin: running the map, keyframes and backend at $384$~pixels with tracking at $256$, together with two fixes to the per-frame pose solve, is a replicated Pareto improvement, $1.28\times$ throughput at $0.964\times$ error on one scene and $1.42\times$ at $0.68\times$ on a second, with $47\%$ less peak GPU memory and $29\%$ less energy per pose. We evaluate on a calibrated, bit-exact regenerable noise ladder, on relabelled real-world dark exposures, and on a new dark-room video ladder recorded from a Boston Dynamics Spot robot.
\end{abstract}

\section{Introduction}
\label{sec:intro}

\begin{figure}[t]
  \centering
  \includegraphics[width=\columnwidth]{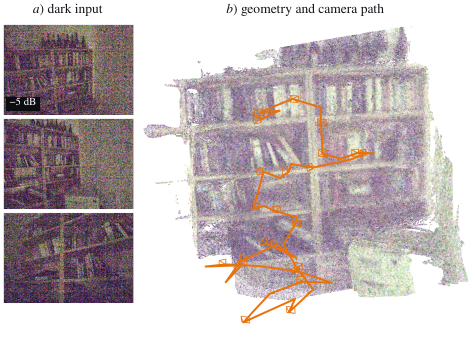}
  \caption{\sys at a glance. \textbf{(a)}~monocular RGB at the photon floor, no
  depth, stereo or IMU, shown with a display gain so the structure is visible;
  \textbf{(b)}~the geometry and camera path recovered from those frames alone,
  on the bookshelf $-5$~dB rung of the calibrated ladder. Every candidate pair
  is gated on the match fraction $\phi$ of Eq.~\ref{eq:matchfrac}: pairs that
  clear the gate become keyframes for a bounded back end, and pairs that do not
  produce \emph{no pose at all}. That gate is why the system stops loudly instead of
  returning a confident trajectory with nothing in it
  (Sec.~\ref{subsec:res-spot}), and why its coverage is the narrowest of the
  methods we compare (Sec.~\ref{subsec:absorbing}).}
  \label{fig:teaser}
\end{figure}

A ground robot repairing electrical problems in a dark basement; an aerial robot flying beneath a forest canopy at dusk to search for a missing hiker; and a ground robot mapping the interior of a collapsed building without light. In all three of these scenarios, the robot must be able to localize and map areas covered and, specifically, may be limited by compute, sensors, and most importantly, light that can be used for vision-based sensing.

The difficulty is the intersection of four constraints, each of which is difficult on its own. The platform is \textbf{compute-constrained}: an Nvidia Jetson Orin AGX draws tens of watts and has a fraction of a workstation GPU's throughput. The scene is \textbf{photon-limited}: with a short exposure forced to avoid motion blur, the signal in a frame is comparable to the sensor's read and shot noise, and the delivered image is dominated with grain rather than by structure. The robot is also \textbf{sensor-constrained}: one rolling-shutter RGB camera, with no LiDAR, no stereo or depth sensor, and no reliable IMU to fall back on when vision fails. And the system must run in \textbf{real time}, emitting a pose while the pose is still useful for control. The four are not independent: a feed-forward front end declares a keyframe whenever it cannot match the current frame against the last one well enough, so darkness raises the keyframe rate, the keyframe rate drives the size of the backend optimisation, and the backend is where the device's time and memory already go (Sec.~\ref{subsec:tractability}).

Each constraint has a mature literature and the pairwise combinations are
well-studied. The four-way intersection is not, and the reason is visible
in what the two relevant families of SLAM actually do. Offline low-light
reconstruction~\cite{dark3r,darkvggt,eag3r,lumos3d,novags} is non-causal,
multi-view and far from a per-frame deadline and many of these methods
require modalities a sensor-constrained robot does not have. Conversely, the real-time
monocular systems a robot can carry, such as DROID-SLAM~\cite{droidslam},
DPV-SLAM~\cite{dpvslam}, VGGT-SLAM~\cite{vggtslam},
$\pi^3$~\cite{pi3}, CUT3R~\cite{cut3r}, EC3R-SLAM~\cite{ec3rslam} and the
classical ORB-SLAM3~\cite{orbslam3} and DSO~\cite{engel2017dso}, all quickly
degrade as the signal-to-noise ratio falls.

We measured this on a calibrated image-power SNR axis with every method ingesting identical frames, and the dominant failure mode is dangerous for when trajectory accuracy is mission-critical. On nine photon-floor rungs, the lowest darkness level of each scene, DROID-SLAM emits a full-coverage trajectory on every rung, and on each, the trajectory carries no more information about the camera's motion than a constant guess at the trajectory centroid. VGGT-SLAM and CUT3R do the same on eight of nine, $\pi^3$ on seven, and DPV-SLAM on four. ORB-SLAM3 and DSO, on the other hand, emit nothing at all. A robot can act on a system that stops; it cannot act on one that confidently provides false trajectories.

\sys (seen in Fig.~\ref{fig:teaser}) is a monocular SLAM pipeline for this setting. It
builds on a feed-forward pointmap front end from the
MASt3R-SLAM~\cite{mast3rslam} lineage, carrying low-light weights, a
photon-aware acquisition policy, and a split-resolution tracking and mapping
path. Our contributions are:

\begin{enumerate}[leftmargin=*,itemsep=1pt,topsep=1pt]
\item \textbf{A measurement of the silent-failure asymmetry, and its
costs.} On a calibrated noise ladder, and on a real-world robot video that
has no priors of the scene, the baselines fail by fabricating
full-coverage trajectories, while our frontend with an explicit match gate does
not. We also show what the gate costs: our coverage is the narrowest of any
method that tracks, because losing tracking in a feed-forward pointmap front
end is an \emph{absorbing state} rather than a transient. We characterize the mechanism and report, from two pre-registered sweeps totaling $208$ runs, that
every configuration-level repair either does nothing or manufactures the same
failure the gate prevents (Sec.~\ref{subsec:absorbing}).

\item \textbf{An embedded execution path using the Nvidia Jetson AGX Orin.} A profile
locates the binding cost in backend global optimization, not the per-frame
network. We implement mpping at $384$~pixels with tracking at $256$, as well as two fixes to the per-frame pose solve, detailed in Sec.~\ref{subsec:deploy}. This improves throughput, error, memory, energy, and tail latency.
\end{enumerate}

\section{Related Work}

\subsubsection{Offline low-light 3D reconstruction}
Dark3R uses a teacher-student distillation approach on Poisson--Gaussian
synthesized raw imagery and reconstructs scenes whose SNR has been reduced
by introducing noise below the $-4$~dB level~\cite{dark3r}. DarkVGGT uses a
thermal channel with physics-aware modeling~\cite{darkvggt}. Both build on
the pointmap paradigm introduced by DUSt3R, which regresses dense pointmaps
directly from pairs of images without requiring known camera calibration or
viewpoint poses~\cite{dust3r}. This family has shown that scene reconstruction
is possible from dark imagery. However, it has not addressed what happens when
a fault appears: the error is smoothed into the SLAM pipeline, producing a
degraded trajectory that is still complete. This is the gap this paper
addresses.

\subsubsection{Real-time monocular SLAM and how darkness affects it}
DROID-SLAM uses bundle adjustment along with dense optical flow as an input
and is the standard learned dense baseline~\cite{droidslam}. Feed-forward
SLAM begins with MASt3R-SLAM~\cite{mast3rslam}, which uses a two-view
pointmap and matching prior to create a dense pointmap. VGGT-SLAM builds on a feed-forward transformer
that infers camera poses, depth maps and point maps in a single pass, and
VGGT-SLAM~2.0 deploys it on a ground robot while aligning submaps on the
$\mathrm{SL}(4)$ manifold~\cite{vggt,vggtslam,vggtslam2}. EC3R-SLAM utilizes the low
latency and low memory footprint of a Jetson Orin~NX~\cite{ec3rslam}. For
contrast, classical feature-based systems such as ORB-SLAM3 remain the
standard non-learned baseline for visual and visual-inertial SLAM across
monocular, stereo, and RGB-D sensors~\cite{orbslam3}. CUT3R~\cite{cut3r}
maintains a single fixed-size persistent state updated at every incoming frame
rather than caching all past frames, and $\pi^3$~\cite{pi3} accepts data in
random order by eliminating reference-view geometry. The gap remains when we want to perform real-time feed-forward SLAM on
photon-limited scenes using a monocular camera.

\subsubsection{Embedded deployment and measurement}
For on-board deployment, EC3R-SLAM utilizes the Orin~NX~\cite{ec3rslam}, and
VGGT-SLAM~2.0 utilizes the Jetson Thor~\cite{vggtslam2}. On the measurement
side, exposure control and auto-exposure benchmarking have been studied for
their effect on visual SLAM~\cite{gamache2025,zhang2017active}, and SNIC
calibrates heteroscedastic noise synthesis against dark frames~\cite{snic2025};
our ladder (Sec.~\ref{subsec:datasets}) is in the same family, labeled by a
measured full-reference image-power SNR in a named linear domain.

\section{Methodology}
\label{sec:method}

\subsection{Pipeline}
\label{subsec:pipeline}

\sys ingests RGB frames from one camera with their capture timestamps and emits
a pose per frame. A feed-forward pointmap network predicts, for a pair of
frames, per-pixel 3D points in a common frame together with per-pixel
confidences and a dense correspondence field, and the tracker solves for the
relative $\mathrm{Sim}(3)$ pose from those correspondences. The backend
maintains a pose graph over keyframes, adding consecutive and
retrieval-proposed wide-baseline edges and refining it by Gauss--Newton. The
network is set with low-light weights distilled by Dark3R~\cite{dark3r}, which is
what makes the front end usable below $0$~dB, and we track at a lower
resolution than we map. The tracker gates
every candidate pair on a \emph{match fraction}: the proportion of pixels whose
correspondence survives a reciprocal-validity mask and confidence tests,
\begin{equation}
\begin{split}
\phi(i,j) \;=\; \frac{1}{|\Omega|}\bigl|\{\,p \in \Omega \;:\;
  & v_p \wedge Q_p > \tau_Q \\[-2pt]
  & \wedge\, C^i_p > \tau_C \wedge C^j_p > \tau_C \,\}\bigr|,
\end{split}
\label{eq:matchfrac}
\end{equation}
where $v_p$ is the reciprocal-match mask, $C$ the per-pixel confidences and
$Q$ the descriptor-confidence score. $\phi$ is the quantity the whole system
turns on: it decides whether a frame is tracked, whether it becomes a keyframe
and, through Sec.~\ref{subsec:absorbing}, whether the run continues at all.

\subsection{Backend bottlenecks}
\label{subsec:profile}

We instrumented the pipeline on a Jetson AGX Orin (JetPack R35.4.1,
\texttt{MAXN}, \texttt{torch}~$2.1$/CUDA~$11.4$, $61$~GB unified) with a probe
recording per-frame service time, per-stage GPU time, backend sub-stage time,
and whole-board power read from the INA3221 rails.

Per-frame service time is bimodal: $19.6\%$ of frames take longer than $0.5$~s
and consume $63\%$ of the wall clock. $81$ of those $98$ frames insert no new keyframes, but the frames that overlap a backend global-optimization pass, with a median service time of $1.350$~s, compared
$0.164$~s for frames that do not, an $8.2\times$ gap. Lock contention, queueing and the interpreter lock are ruled
out by measurement, leaving compute contention on the default CUDA stream, and
the two paths are almost perfectly additive: front end alone $88.6$~s, backend
busy $121.2$~s, together $198.6$~s, $95\%$ of the sum.

Within the backend, $59\%$ of the time is the symmetric decode of candidate
edges at the map resolution, whether or not the gate later accepts them, and
$37\%$ is the Gauss--Newton solve, which grows $6.4\times$ over a run as the
graph grows. Hence, we show that the map resolution, not the
tracking resolution, is the expensive knob because it sets the cost of the
dominant decode. Second, $8$--$10$ poses/s is arithmetically unreachable with global optimization in the loop: the front end alone runs at $5.65$ poses/s, the backend alone needs $84$--$121$~s per $500$ frames, and the two are additive, so even a zero-cost
front end caps out at $6.0$ poses/s. The fastest point we measured is $7.35$
poses/s with the backend deleted, at $47.65\%$ of the no-information ceiling,
$2.2\times$ the error.

\subsection{The deployable configuration}
\label{subsec:deploy}

\begin{table}[t]
\centering
\vspace*{5pt}
\captionsetup{font=normalfont, labelfont=bf}
\caption{The deployable configuration against the previous deployment, chapel
$-10$~dB (\textbf{synthetic}, tripod stills), Jetson AGX Orin, \texttt{MAXN}.
Three runs per arm; the two distributions are disjoint on both the throughput
and the accuracy axis, with three-run ranges under $3\%$. ATE is the all-frame
trajectory as \% of the no-information ceiling.}
\label{tab:pareto}
\small
\begin{tabular}{lccc}
\toprule
 & previous & \textbf{\sys} & ratio \\
\midrule
poses/s                & $2.512$ & $\mathbf{3.204}$ & $\mathbf{1.28\times}$ \\
ATE (\% of ceiling)    & $21.47$ & $\mathbf{20.70}$ & $0.964\times$ \\
peak CUDA (GiB)        & $20.70$ & $\mathbf{10.95}$ & $-47\%$ \\
energy per pose (J)    & $14.84$ & $\mathbf{10.47}$ & $-29\%$ \\
latency p95 (s)        & $1.384$ & $\mathbf{0.907}$ & $-34\%$ \\
\bottomrule
\end{tabular}
\vspace{-1.2em}
\end{table}

\sys drops the map, keyframe, relocalization and backend resolution
from $512$ to $384$ while tracking stays at $256$, reducing the cost of every
backend decode without touching the tracker's correspondence field. We
then applied two changes to the calibrated pose solve, which the profile
at $24.5\%$ of front-end time: instead of building a full-resulution Jacobian every run, a configured residual subsampling was added, and a
scalar convergence test costing two device synchronizations per iteration is
replaced by a device-side test costing one.

The result is a Pareto improvement, not a trade (Table~\ref{tab:pareto}). Over
three runs per arm on chapel at $-10$~dB, throughput rises $1.28\times$ ($2.512$ to $3.204$ poses/s)
while error \emph{falls} to $0.964\times$ ($21.47$ to $20.70\%$ of ceiling),
with peak GPU memory down $47\%$ ($20.70$ to $10.95$~GiB), energy per pose down
$29\%$ ($14.84$ to $10.47$~J) and p95 latency down $34\%$ ($1.384$ to
$0.907$~s); the two three-run distributions are disjoint on both axes, with
three-run ranges under $3\%$. A second scene reproduces and enlarges it: on
bookshelf at $-10$~dB, over two runs per arm, the same configuration reaches
$1.42\times$ the throughput at $0.68\times$ the error.

\subsection{Darkness degrades tractability, not only accuracy}
\label{subsec:tractability}

The memory figure matters as much as the throughput figure, because of a
coupling between two of the four constraints. A feed-forward front end declares
a keyframe when it cannot match the current frame well enough against the last
one, so the keyframe rate rises as the light falls: on the same scene
it is $13\%$ of frames at $-10$~dB and $54\%$ at $-15$~dB. The pose graph grows
with the keyframe count, with the CUDA caching allocator's pool because
each backend pass requests buffers sized by the current graph and can never
reuse a cached block. Host memory stays flat at $8.5$~GB of resident set while
the allocator pool climbs to $45.0$~GiB. The consequence is not slow operation
but no operation: on a $500$-frame chapel $-15$~dB rung the previously deployed
configuration exhausts all $61$~GB and is killed, deterministically, at frames
$350$ and $349$ of two runs. A guard does not help because the optimization-window bound exempts loop-closure
edges and almost every edge is one; a strict bound halves peak memory but costs
$1.9$--$2.9\times$ the error. Reducing the map resolution attacks the same
growth from the other side and produces the $47\%$ memory saving.

\section{Experimental Setup}
\label{sec:setup}

\subsection{Measurement contract}
\label{subsec:contract}

Darkness is indexed by a frozen full-reference image-power SNR. With $S$ a
clean reference frame and $Y$ the delivered frame of the same viewpoint, both
in the linear, black-subtracted sensor domain,
\begin{equation}
\mathrm{SNR} \;=\; 10 \log_{10}
\frac{\sum_{\mathcal{M}} S^{2}}{\sum_{\mathcal{M}} (Y - S)^{2}},
\label{eq:snr}
\end{equation}
with energies summed over $R$, $G$, $B$ and over the frame before the
logarithm, $\mathcal{M}$ every pixel and channel, and the black level fitted
per scene and channel rather than assumed. Three stages are measured and are
not interchangeable, because the gamma-domain sRGB value sits $0$--$18$~dB above the
linear one and the gap grows as the light falls. All dB figures here are
pre-clip linear unless stated.

Re-measuring the $23$ real dark
exposures of the Dark3R ladder on Eq.~\ref{eq:snr} moves them from $-18.40$ to
$-0.43$~dB on the retired per-channel amplitude ratio to $-17.59$ to
$+5.46$~dB (Spearman $0.73$). The label is itself a lower bound, since parallax and illumination residuals between two hand-held walks remain in the error energy.

\subsection{Datasets}
\label{subsec:datasets}

\paragraph{Calibrated synthetic ladder}
Fifteen rungs are synthesized from the brightest real exposure of three scenes
(chapel, bookshelf, traincar) at pre-clip targets $\{0,-5,-10,-15,-20\}$~dB
under a calibrated Poisson--Gaussian sensor model. The largest realized miss on
any rung
is $0.001$~dB against a $0.3$~dB tolerance, and every rung regenerates
bit-exactly from a single seed. The reference $S$ is itself noisy, so the
synthesis subtracts its own scaled noise, making total noise relative to the
true scene match a real capture at that exposure. Ground truth is COLMAP at arbitrary scale, so errors on this
ladder are in COLMAP units and never called meters. These are tripod stills,
consumed every eighth frame.

\paragraph{Spot dark-room video ladder}
\begin{table}[t]
\centering
\vspace*{5pt}
\captionsetup{font=normalfont, labelfont=bf}
\caption{Our Dark-room video ladder. Median $|\nabla I|$ is the spatial gradient a matcher consumes. Rung 0 is a null-input control: $86.5\%$ of its delivered frames are identically zero.
These are H.264 with no calibrated SNR.}
\label{tab:spot}
\small
\begin{tabular}{llrrrr}
\toprule
rung & mean luma & frac $<5$ & med.\ $|\nabla I|$ & levels \\
\midrule
0 (lights on) & $55.39$ & $8.3\%$  & $0.690$ & $169$ \\
1 & $5.99$ & $58.0\%$ & $0.049$ & $83$ \\
2 & $1.69$ & $95.6\%$ & $0.020$ & $69$ \\
3 & $0.68$ & $98.8\%$ & $0.005$ & $8$ \\
4 (null) & $0.23$ & $99.2\%$ & $\mathbf{0.000}$ & $\mathbf{1}$ \\
\bottomrule
\end{tabular}
\vspace{-1.2em}
\end{table}

\begin{figure}[b]
  \centering
  \begin{minipage}[t]{0.49\columnwidth}
    \centering
    \includegraphics[width=\linewidth]{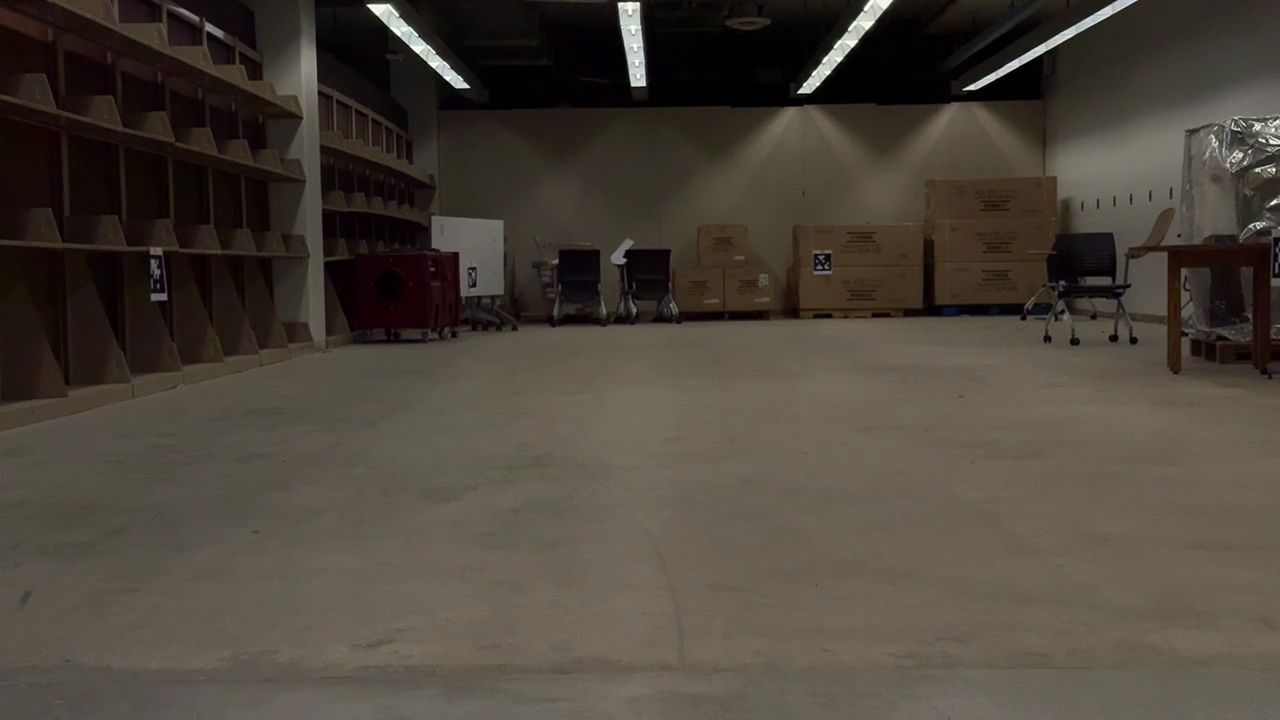}\\[-0.2em]
    {\footnotesize (a) lights on, rung 0}
  \end{minipage}\hfill
  \begin{minipage}[t]{0.49\columnwidth}
    \centering
    \includegraphics[width=\linewidth]{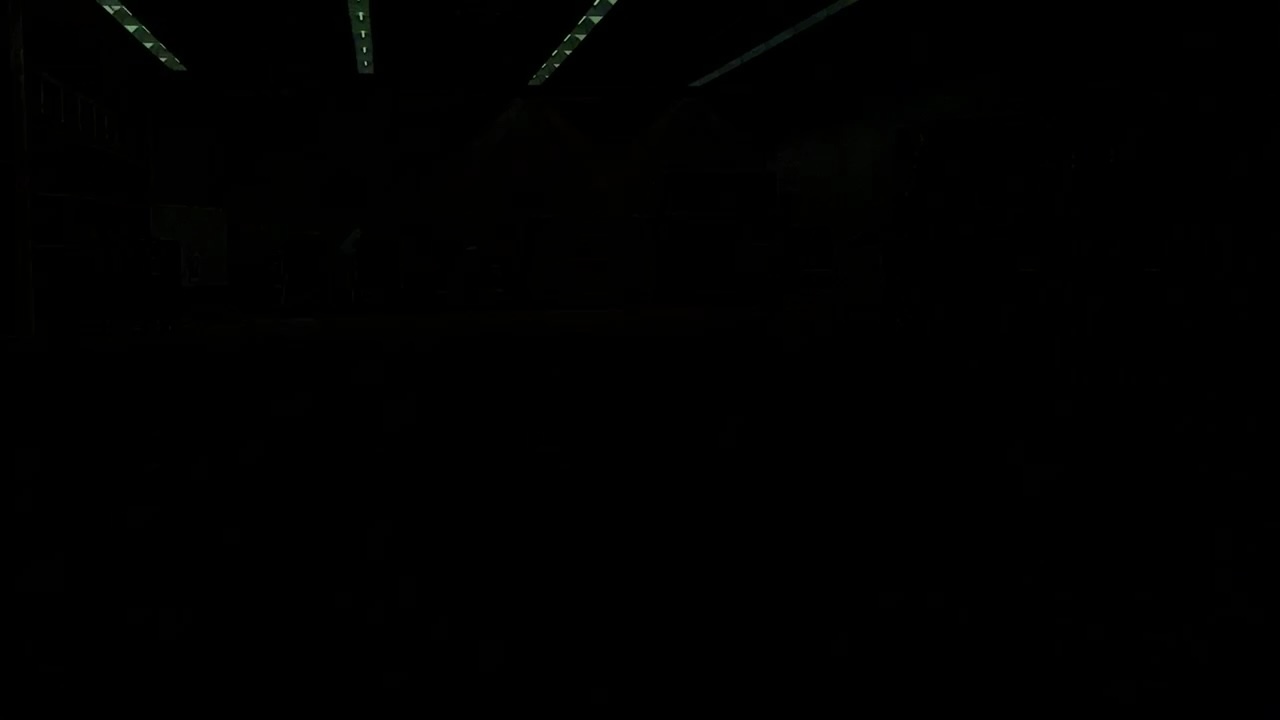}\\[-0.2em]
    {\footnotesize (b) lights off, dark rung}
  \end{minipage}
  \caption{The Spot dark-room video ladder scene as delivered by the
  iPhone~15~Pro carried on the robot, with no display gain applied.
  \textbf{(a)}~the lights-on control take (rung 0 of Table~\ref{tab:spot});
  \textbf{(b)}~the same warehouse on a dark rung, where only the ceiling
  fixtures remain above the codec's black level.}
  \label{fig:spot-room}
\end{figure}

Five takes of a Boston Dynamics Spot walking one nominal zigzag path across a dark
warehouse and back, filmed from an iPhone~15~Pro carried on the robot at fixed
ISO~$400$ with varied exposure (Fig.~\ref{fig:spot-room}). This is our real \emph{continuous-video}
low-light experiment as earlier dark rungs were tripod stills.
Table~\ref{tab:spot} characterizes it: the median spatial gradient falls by more
than four orders of magnitude from the lit take to the darkest. Three
properties bound what the data can support. The robot walks out and
back, so $\|p_{\mathrm{last}}-p_{\mathrm{first}}\|$ over path length is
scale-invariant and reference-free. There is no calibrated SNR: these are H.264 deliveries at $3.51$~Mbps down to
$51$~kbps, and in the two darkest takes $96$--$99\%$ of macroblocks carry a
single repeated value, so the codec has quantized away the denominator of
Eq.~\ref{eq:snr} and the takes are ordered by exposure rather than placed on
the dB axis. The darkest take is retained deliberately as a
\textbf{null-input control}: a method that emits a confident, structured
trajectory on it is fabricated. Every method receives identical
frames at a uniform stride of two, on the container timestamps rather than an
assumed grid, with the baselines re-run at native rate as a
decimation-robustness control.

\paragraph{Supporting datasets}
The $23$ relabeled real dark exposures of Sec.~\ref{subsec:contract} are
tripod stills with COLMAP ground truth, and the only real data we have on the
contract below $-8$~dB. \emph{CAVERS} and \emph{EVIMO2} (four segments of one
recording, mocap ground truth) are the native-rate held-out tests, with the
verdict taken on the pooled row, and \emph{TUM fr1} with calibrated darkening
is the bright/dark paired control.

\subsection{Protocol, baselines and metrics}
\label{subsec:protocol}

Every method receives the same input: RGB frames of one camera at native rate
with timestamps, plus intrinsics where the method accepts them. No IMU, depth,
events, stereo, long exposure, ground-truth pose or scale reaches any method,
even where the dataset includes it. Baselines run in monocular mode at upstream
defaults, are verified on a bright reference before any dark failure is
interpreted, and use one configuration across every sequence of a table,
namely DROID-SLAM~\cite{droidslam}, DPV-SLAM~\cite{dpvslam},
VGGT-SLAM~\cite{vggtslam}, $\pi^3$~\cite{pi3}, CUT3R~\cite{cut3r},
EC3R-SLAM~\cite{ec3rslam}, ORB-SLAM3~\cite{orbslam3} and
DSO~\cite{engel2017dso}; VGGT-SLAM and EC3R-SLAM predict their own intrinsics
and are run uncalibrated.

Accuracy is $\mathrm{Sim}(3)$-aligned ATE RMSE, reported for arbitrary-scale
ground truth as a percentage of the \emph{no-information ceiling}, the RMS
radius of the ground-truth positions about their centroid; a method at $100\%$
of ceiling is no better than a constant guess at the centroid. Two supports are
always labeled: common support (\comsup), the timestamps present in every
compared arm, and native support (\natsup), the poses a method actually
emitted. Where coverage differs greatly \comsup{} differentiates the better-covered
method, so classification uses \natsup. A run is \emph{tracked} if it emits at
least $160$ poses at native rate ($20$ under the every-eighth stills protocol)
and its error is below $75\%$ of ceiling, a \emph{silent failure} if it fails
that test at $\ge 75\%$ coverage, and a \emph{loud failure} otherwise. Every
attempted run appears.

\section{Results}
\label{sec:results}

\subsection{Behaviour at the photon floor}
\label{subsec:res-floor}

\begin{table}[t]
\centering
\vspace*{5pt}
\captionsetup{font=normalfont, labelfont=bf}
\caption{Tracked / silent / loud over the nine synthetic floor rungs
($-10/-15/-20$~dB $\times$ three scenes); every-8
stills protocol.}
\label{tab:silent}
\small
\begin{tabular}{lccc}
\toprule
method & tracked & silent & loud \\
\midrule
EC3R-SLAM~\cite{ec3rslam}       & \textbf{5} & 4 & 0 \\
DPV-SLAM~\cite{dpvslam}         & 4 & 4 & 1 \\
\textbf{\sys, full}             & 3 & \textbf{0} & 6 \\
$\pi^3$~\cite{pi3}              & 2 & 7 & 0 \\
CUT3R~\cite{cut3r}              & 1 & 8 & 0 \\
VGGT-SLAM~\cite{vggtslam}       & 1 & 8 & 0 \\
\sys, $k{=}1$ front end         & 1 & \textbf{0} & 8 \\
DROID-SLAM~\cite{droidslam}     & 0 & \textbf{9} & 0 \\
ORB-SLAM3 / DSO                 & 0 & 0 & 9 \\
\bottomrule
\end{tabular}
\vspace{-1.2em}
\end{table}

Three properties of Table~\ref{tab:silent} matter for a robot. \sys returns no
uninformative full-coverage trajectory on any rung of this protocol, in any of
the $74$ runs we made, including two arms built specifically to widen coverage
by feature averaging. Where it tracks it is the most accurate method by a
factor of two to four, at $24$--$48\%$ of ceiling against EC3R-SLAM's
$56$--$73\%$ and DPV-SLAM's $57$--$68\%$. And that accuracy is a result of the
narrowest coverage of any method that tracks, as explained in 
Sec.~\ref{subsec:absorbing}.

Of the $21$ floor cells tracked on common
support, $18$ are within about $5$ percentage points on native support and
three flip to silent: EC3R-SLAM on chapel $-15$~dB, which reads $35\%$ of
ceiling on the $19$ poses we also cover while its own $61$-pose trajectory
scores $92.4\%$; DROID-SLAM on the same rung ($60$ to $86\%$); and CUT3R on
bookshelf $-10$~dB ($67$ to $79.5\%$). Our own counts are identical under both
conventions, and the classification on chapel $-10$~dB was
independently re-derived from the raw trajectories, where all five learned
methods emit exactly $63$ poses and four of them score between $84\%$ and
$99\%$ of a ceiling of $3.6585$ COLMAP units against $23.95\%$ for \sys.

\subsection{Native-rate video}
\label{subsec:res-native}

Accuracy is protocol-specific and reverses on continuous video. On
EVIMO2, pooled over the four segments of the one recording at full coverage for
every method, DROID-SLAM reaches $5\%$ of ceiling, DPV-SLAM $8\%$, our
$k{=}1$ (no accumulation) front end $9\%$ and VGGT-SLAM $15\%$. The geometric baselines lead, and our own $k{=}1$ configuration beats our tuned deployment
configuration, which reaches $15.6\%$ on segment~1 against $7.5\%$ for $k{=}1$
on the same segment. Split-resolution tracking accounts for about half of the
gap, and it is not exercised by the stills protocol, where every input frame
becomes a keyframe; on video roughly one frame in nine is a keyframe, so most
reported poses are raw fits to a keyframe that is never revisited. We therefore
report $k{=}1$ as the configuration for continuous video, and the tuned
configuration for the stills and device results.

\subsection{Real robot video: the null-input control}
\label{subsec:res-spot}

\begin{table}[t]
\centering
\vspace*{5pt}
\captionsetup{font=normalfont, labelfont=bf}
\caption{The null-input control, $86.5\%$ of whose delivered frames are
identically black: the $1178$ frames of the stride-two matrix, every method on
identical input. Closure is return-to-start error over path length. No calibrated SNR.}
\label{tab:null}
\footnotesize
\setlength{\tabcolsep}{4pt}
\begin{tabular}{lcccl}
\toprule
method & poses & cov. & closure & outcome \\
\midrule
\sys, $k{=}1$ & $20$   & $0.017$ & $0.972$ & stops after preamble \\
\sys, full    & $20$   & $0.017$ & $0.905$ & stops after preamble \\
\sys, deploy  & $22$   & $0.019$ & $0.999$ & stops; $1149$ reloc fails \\
DROID-SLAM    & $\mathbf{1178}$ & $1.000$ & $\mathbf{0.040}$ & \textbf{silent} \\
DPV-SLAM      & $\mathbf{1178}$ & $1.000$ & $0.354$ & \textbf{silent} \\
VGGT-SLAM     & $537$  & $0.456$ & n/a     & crash: degenerate rot. \\
EC3R-SLAM     & $0$    & $0.000$ & n/a     & hung; out of memory \\
\bottomrule
\end{tabular}
\vspace{-1.2em}
\end{table}

The clear objection to Table~\ref{tab:silent} is that its rungs are
synthetic. The Spot ladder shows results on real robot video, and its sharpest
probe is the null-input control, a $1.27$~s lit preamble followed by $77.2$~s below $0.5$~DN. Three
independent estimators agree the remainder carries no information: $86.5\%$ of
delivered frames are identically zero at a standard deviation of exactly
$0.000$; the median spatial gradient is $0.000$; and the pointmap network's own
focal estimate degenerates to $26{,}457$~px where the four takes carrying
signal give $808$--$970$~px. Any structured trajectory on this take is
fabricated.

All three \sys configurations emit exactly the lit preamble (the beginning of the video had lights on for a couple seconds) and then stop
(Table~\ref{tab:null}); the deployed one attempts $1149$ relocalisations first,
so the abstention is worked for rather than incidental. DROID-SLAM and DPV-SLAM
return smooth, structured trajectories from an input that carries zero
bits. DROID-SLAM's return-to-start
closure on the null input is $0.040$, better than its closure on the lights-on
take of the same room ($0.095$). A fabricated trajectory scores
better on closure than one estimated from real imagery.

\subsection{On-device throughput, latency and power}
\label{subsec:res-orin}

\begin{table*}[t]
\centering
\vspace*{5pt}
\captionsetup{font=normalfont, labelfont=bf}
\caption{Jetson AGX Orin, \texttt{MAXN}, chapel $-10$~dB (\textbf{synthetic},
tripod stills). Latency is acquisition-to-pose with a free-running source.
Power is the whole-board INA3221 sum, mean over the run; ATE is the all-frame
trajectory as \% of the no-information ceiling. The first three rows are single
runs; the last is the configuration of Sec.~\ref{subsec:deploy}, three-run
mean.}
\label{tab:orin}
\small
\begin{tabular}{lccccc}
\toprule
configuration & poses/s & p50/p95/p99 (s) & W & J/pose & ATE \\
\midrule
every-8th, colour $512$ & $0.411$--$0.413$ & $0.416$/$0.518$/$1.54$ & $34.8$--$35.2$ & $87$ & $22.75\%$ \\
every-frame, split $256$/$512$ & $2.54$ & $0.166$/$1.36$/$1.85$ & $36.2$ & $14.8$ & $21.47\%$ \\
every-frame, floor mode & $2.485$ & $0.312$/$1.17$/$1.74$ & $39.6$ & $16.1$ & $40.1\%$ \\
\midrule
\textbf{\sys (map $384$ + solver fixes)} & $\mathbf{3.204}$ & $\mathbf{0.163}$/$0.907$/$1.17$ & $\mathbf{32.6}$ & $\mathbf{10.5}$ & $\mathbf{20.70\%}$ \\
\bottomrule
\end{tabular}
\vspace{-1.2em}
\end{table*}

Table~\ref{tab:orin} gives the device measurement. Moving from the every-eighth
protocol to every-frame split-resolution operation is worth eight times the
poses at a sixth of the energy per pose for slightly
better accuracy on common support ($20.94\%$ against $22.75\%$), and the
configuration of Sec.~\ref{subsec:deploy} improves on that on every axis at
once. Every run emitted a pose for every input frame, with no thermal
throttling and a peak junction temperature of $63.6^{\circ}$C, and throughput,
power and memory reproduce to about $1\%$ between runs.

\section{Analysis}
\label{sec:analysis}

\subsection{One nominal SNR is not one task difficulty}
\label{subsec:difficulty}

A calibrated axis makes rungs comparable, not equally hard.
At a common $-15$~dB level the median translation-direction error of a
single-pair baseline over a frozen endpoint set runs from $25.2^{\circ}$ on TUM
\texttt{fr1/desk} to the $90^{\circ}$ chance ceiling on CAVERS and EVIMO2, the
entire dynamic range of the metric at one label. Part of the cause is that
EVIMO2 is quantization-limited before it is even darkened, with a mean of $3$~DN and
only $31$--$77$ distinct intensity levels per frame. The wide-baseline edge
gate likewise accepts $74\%$ of $148$ candidates on chapel at $-10$~dB but
only $16\%$ of $421$ on bookshelf at the same level, and an offline non-causal
reconstructor carries no trajectory information below $-15$~dB on chapel,
$-10$~dB on bookshelf and $-5$~dB on traincar, a $10$~dB spread.
</br>

One property of that offline upper bound bears directly on how a deployed
system should be built: the objective cannot rank its own
reconstructions. The solver's
final-loss ratio relates to the true error ratio at Pearson $r = +0.411$
($p = 0.27$) with $13$ of $36$ comparisons misordered, and every pair in which
both arms exceed $75\%$ of ceiling reports a loss ratio between $1.01$ and
$1.14$ while its error ratio is meaningless. The optimizer reports near-parity
exactly when both reconstructions have collapsed, which is why a competence
signal has to come from outside the objective.

\subsection{Tracking loss is an absorbing state, and no configuration escapes it}
\label{subsec:absorbing}

The coverage column of Table~\ref{tab:silent}, three rungs tracked against
EC3R-SLAM's five, has a single, identifiable cause, and it is neither the
imagery nor the matcher. When $\phi$ falls below the tracking gate ($0.05$) the frame is declared lost
and the tracker enters relocalization, which scores it against retrieved
keyframes under a gate six times stricter ($0.30$), evaluated against a map
that early in a sequence may hold one node. Escape therefore requires a pair
six times better than the one that just failed, against fewer candidates than
were available before. On chapel at $-10$~dB the chain contains exactly one
sub-gate link, at $\phi = 0.0499$, and that link costs $43$ of $63$ poses. On
traincar at $0$~dB, $42$ of the $49$ consecutive links clear the tracking gate
and the first misses it at $0.036$, but \emph{no pair anywhere in the sequence}
reaches $0.30$ against keyframe~$0$, the measured maximum being $0.1805$, so
all $83$ relocalization attempts fail and the run ends at one pose. On the same
frames, an offline non-causal reconstructor using bit-identical weight tensors
reaches $17.4\%$ of ceiling, and our matcher ranks the pairs it uses at
Spearman $+0.904$ against its correspondence counts. The imagery is not the
problem and the matcher is not the problem; the control flow is.

A seeded-re-matching study rules out the obvious repair. Re-seeding the
correspondence search lifted $0$ of $279$ failure events above the
gate, including under a ground-truth-pose oracle; the search is already
effectively global, so seeding was never the constraint. What removes the
correspondences is the descriptor-confidence term $Q_p$ in
Eq.~\ref{eq:matchfrac}: on $268$ of the $279$ events the validity mask still
covers $18$--$53\%$ of pixels while the median survival rate through the
confidence test is $0.0000$. The front end discards correspondences it has
already found.

\begin{table}[t]
\centering
\vspace*{5pt}
\captionsetup{font=normalfont, labelfont=bf}
\caption{Config-only escape routes from the absorbing state, over six
development floor rungs (\textbf{synthetic}, every-eighth stills), each
pre-registered before its runs and fitted on development scenes only. Coverage
is pooled over the six rungs. \textbf{No lever clears the pre-registered
endpoint}, which required two additional tracked rungs with no new silent
failures.}
\label{tab:levers}
\small
\begin{tabular}{lccc}
\toprule
lever & coverage & tracked & silent \\
\midrule
as shipped                                   & $16.5\%$ & $1$ & $0$ \\
relocalization gate $0.30 \!\to\! 0.05$      & $16.5\%$ & $1$ & $0$ \\
tracking gate $0.05 \!\to\! 0.01$            & $17.5\%$ & $1$ & $0$ \\
bounded map re-initialization                & $41$--$62.5\%$ & $1$--$2$ & $0$ \\
\bottomrule
\end{tabular}
\vspace{-1.2em}
\end{table}

The clear remedy is to relax the gate, or its asymmetric recovery path, into
a margin. On chapel at $-10$~dB that works: coverage rises from $32\%$ to
$100\%$ for a $34\%$ relative increase in trajectory error, an operating point
a robot would take. It does not survive a held-out test.
Table~\ref{tab:levers} reports every configuration-level escape route we
pre-registered and fitted on a development split of two scenes, over $93$
runs, with the unmodified arm
reproducing the published floor row of Table~\ref{tab:silent}. Closing the
$6\times$ asymmetry from the principled side is exactly inert: taking the
relocalization gate from $0.30$ to $0.05$ reproduces the baseline cell for cell
on all six development rungs. Lowering the tracking gate is nearly
inert: its entire effect is $+3$ poses on one rung, and the value that worked
on chapel yields one pose on five of six development rungs, so that value was a
single-scene coincidence, which is exactly what the held-out split existed
to catch. Map re-initialization is the only lever that moves coverage,
and on the hardest scene it moves it into the no-information band: coverage
rises from $2\%$ to $54\%$ on traincar with every scored cell between $78.5\%$
and $95.1\%$ of ceiling, and merging the resulting components makes the
trajectory worse than leaving them apart. The pre-registered joint endpoint is
formally undetermined, because no arm reached the held-out scene before the
runs were stopped, but two of its six kill criteria fire on development data
alone. One caveat belongs with the zero in the silent column: across those $93$
runs no lever produced a silent failure, but eight cells sit above $75\%$ of
ceiling and are classified loud only because coverage stayed below $0.75$, one
of them by $8.3$ percentage points. The record survived partly because the
levers were too weak to do damage.
The one lever strong enough does damage, and only the three-way rule
sees it. A second pre-registered sweep, $115$ runs, took the
descriptor-confidence threshold $\tau_Q$ from its shipped value down to zero
across all nine floor rungs. It is the right knob: at $0.25$ pooled floor
coverage rises from $27.8\%$ to $96.2\%$ and the tracked count from three to
six, the largest coverage movement we measured. Every setting that
produces it also produces two to four full-coverage trajectories between $75\%$
and $100\%$ of the no-information ceiling. Traincar at $-10$~dB goes from one
pose and a loud failure to $50$ of $50$ poses with no relocalisation failures,
at $99.6\%$ of ceiling; at $-15$~dB one setting scores $95.3\%$, worse than
DROID-SLAM's own silent failure on that rung ($83.8\%$), and the fabrication is
not confined to the floor: it appears at $0$ and $-5$~dB as well. This is a
plateau, not a tuning accident: silent failures appear at every one of the six
settings from $0.5$ down to $0$, on all three scenes, and no run of settings
passes the three-way rule. The accuracy criterion passes on every arm,
because error on the rungs that were already tracked stays fine; a
coverage-only or accuracy-only reading of this experiment would have shipped a
fabricating system. Nor is the gate discriminating arbitrarily: forcing the
solver onto rejected links gives median rotation error ordered by gate group on
eight of nine rungs (chapel $-15$~dB: $5.9^{\circ}$, $6.8^{\circ}$,
$33.9^{\circ}$), with match fraction and rotation error negatively correlated
on all nine. The gate is doing real work, and what it buys is the absence of a
trajectory when the links are hopeless.

We therefore report the absorbing state as a measured failure mechanism, not a threshold to be tuned away.
\subsection{Other pre-registered negatives}
\label{subsec:negatives}

Four further mechanisms were pre-registered with a gate and a kill clause and
are reported for the design directions they close.
\emph{Retained-frame recovery}, which attaches the historical factor a lost
frame would otherwise drop, was killed: the scale gap moves the wrong way,
$81.3$ to $91.5$. \emph{Temporal evidence accumulation} missed its bar, with one
of three evaluable recordings clearing the pre-registered $97.5\%$
threshold. \emph{Coverage by feature averaging with abstention}
failed on both halves: coverage rose on zero of nine rungs, and the abstention
rule was inert because our failures at the floor are already loud. And a
\emph{competence monitor}, meant to predict silent failure in methods with no
gate of their own, failed all three of its gates: on the floor rungs its AUROC
is $0.500$, exactly no information, at $60$--$100\times$ the cost of a $1.7$~ms
photon proxy that beats it.

\subsection{Limitations}
\label{subsec:limitations}

\textbf{Calibrated evidence depth.} Every calibrated result below about
$-8$~dB is synthetic, from three scenes and one sensor model; the only real
exposures we have on the contract below that level span $-8.14$ to $-15.22$~dB.
The Spot ladder is real and continuous but compressed, with no ground truth and
no dB label: one room, one robot, one session, $n=1$ per rung.

\textbf{The failure-mode classification is protocol-dependent.} Under the
\emph{every-frame} protocol four of $44$ classifiable device runs are silent
failures by our own rule: the deployable configuration on chapel $-15$~dB twice
($83.4\%$ and $83.1\%$ of ceiling), the same with no backend at all ($81.2\%$),
and a raised match-fraction threshold on the reference rung ($94.8\%$). No configuration should be offered as a deployment default without
that qualification. Whether the deployable configuration \emph{causes} these or
merely \emph{reveals} them is open: the previous map-$512$ configuration never
produced a classifiable trajectory on that rung, because it exhausts the device
first (Sec.~\ref{subsec:tractability}), so our reading that the effect is
protocol-driven rests on the no-backend arm landing in the same band, an
inference rather than a direct comparison.

\textbf{Throughput.} \sys reaches $3.204$ poses/s on the Orin: real-time for a
slow ground robot, not for a $30$~Hz video stream, and
Sec.~\ref{subsec:profile} shows the gap cannot be closed by tuning with global
optimisation in the loop.

\section{Future Work}
\label{sec:future}

The natural successor to the diagnosis of Sec.~\ref{subsec:absorbing} is to
\textbf{correct the trajectory rather than only refuse to extend it}. Admitting
a degraded constraint converts an absorbing stop into a degraded but continuing
estimate; it does not  tell us \emph{which}
constraints were degraded and by how much in the system. Every admitted pair carries its own
match fraction, and Sec.~\ref{subsec:absorbing} shows that quantity is well
correlated with what an offline matcher finds, so the pose graph could weight a
degraded edge by its margin, mark the interval it spans as provisional, and
revisit it when a later wide-baseline retrieval edge re-observes the same
structure at a higher match fraction, repairing the gauge of the provisional
segment instead of carrying its error forward. This is a backend change, and
Sec.~\ref{subsec:profile} shows the backend is where the compute already goes,
so it must be designed against the same bounded-window budget.

Three further directions follow from the measurements. The optimiser's
inability to rank its own reconstructions means a competence signal must come
from outside the objective; the off-the-shelf-matcher route is closed, but a
signal read from the front end's own match-fraction distribution over a window,
which costs nothing extra, is not. The throughput gap is in backend global
optimisation, so an incremental or submap-local solver, rather than a faster
per-frame network, is the lever. And the evidence base needs a real capture
below $-15$~dB on the contract: the Spot ladder shows that a consumer codec
destroys the noise structure the contract measures, so such a capture has to be
taken in raw.

\section{Conclusion}
\label{sec:conclusion}

We presented \sys, a monocular SLAM pipeline for the intersection of four
constraints a dark-environment robot faces at once. Across nine photon-floor
rungs it returns no uninformative full-coverage trajectory, where DROID-SLAM
returns nine, and where it tracks it reaches the lowest error of any method on
that protocol; on a robot video take whose frames are $86.5\%$ identically
zero, all three of our configurations stop after the lit preamble while two
baselines emit a pose for all $1178$ frames. That property comes at the cost of the narrowest coverage of any method that
tracks, and we show why: tracking loss is an absorbing state, and $208$
pre-registered runs establish that no configuration-level repair generalises.
The one knob strong enough to widen coverage does so by fabricating
trajectories that only a three-way tracked/silent/loud rule detects. On the device, we located the binding cost in backend global optimisation
rather than in the per-frame network, and gave a configuration that improves
throughput, error, memory, energy and tail latency at once, on two scenes.

\section*{Acknowledgment}
We acknowledge the use of Claude 5.1 Fable to improve grammar, clarity, phrasing, and overall language refinement. The core ideas, technical contributions, and data remain entirely the work of the authors, who accept responsibility for the manuscript.

\bibliographystyle{IEEEtran}
\bibliography{references}

\end{document}

%% file: preamble.tex
\usepackage[T1]{fontenc}
\usepackage{times}
\usepackage{cite}
\usepackage{amsmath,amssymb,amsfonts}
\usepackage{mathtools}
\usepackage{graphicx}
\usepackage{textcomp}
\usepackage{xcolor}
\usepackage{booktabs}
\usepackage{multirow}
\usepackage{array}
\makeatletter\let\labelindent\undefined\makeatother
\usepackage{enumitem}
\usepackage{url}

\makeatletter
\let\NAT@parse\undefined
\makeatother
\usepackage[colorlinks=true,linkcolor=blue,citecolor=red,urlcolor=cyan,
            pdftitle={Noctif3R}]{hyperref}

\graphicspath{{figures/}}

\usepackage{xspace}
\newcommand{\sys}{\textsc{Noctif3R}\xspace}

\newcommand{\natsup}{\textsc{nat}}
\newcommand{\comsup}{\textsc{com}}